\documentclass{article}

\usepackage{arxiv}

\usepackage{hyperref}

\usepackage{graphicx}
\usepackage{float}
\usepackage{amsfonts}
\usepackage{mathtools}
\usepackage{amsmath}
\usepackage{amssymb}
\usepackage{todonotes}
\usepackage{bbold}
\usepackage{amsthm}
\usepackage{multirow}
\usepackage{makecell}
\usepackage{booktabs}
\usepackage{xspace}
\usepackage{txfonts}
\usepackage[T1]{fontenc}

\newtheorem{property}{Property}[section]

\newtheorem{theorem}{Theorem}
\newtheorem{definition}[theorem]{Definition}

\newcommand{\player}{\textit{Player}\xspace}
\newcommand{\players}{\textit{Players}\xspace}
\newcommand{\round}{\textit{Round}\xspace}
\newcommand{\rounds}{\textit{Rounds}\xspace}
\newcommand{\score}{\textit{Score}\xspace}
\newcommand{\scores}{\textit{Scores}\xspace}
\newcommand{\leaderboard}{\textit{Leaderboard}\xspace}
\newcommand{\leaderboards}{\textit{Leaderboards}\xspace}
\newcommand{\tournament}{\textit{Tournament}\xspace}
\newcommand{\tournaments}{\textit{Tournaments}\xspace}
\newcommand{\scheme}{\textit{Scheme}\xspace}
\newcommand{\schemes}{\textit{Schemes}\xspace}
\newcommand{\opponent}{\textit{Opponent}\xspace}
\newcommand{\opponents}{\textit{Opponents}\xspace}
\newcommand{\metascore}{\textit{Meta-Score}\xspace}
\newcommand{\metascores}{\textit{Meta-Scores}\xspace}
\newcommand{\match}{\textit{Match}\xspace}
\newcommand{\matches}{\textit{Matches}\xspace}

\usepackage{natbib}
\begin{document}

\title{Interpretable Meta-Score for Model Performance}

\author{Alicja Gosiewska \\
        \texttt{ alicjagosiewska@gmail.com }\\
        Why R? Foundation
    \And
       Katarzyna Wo\'{z}nica$^*$ \\
       \texttt{katarzyna.woznica.dokt@pw.edu.pl} \\
       Faculty of Mathematics and Information Science\\
       Warsaw University of Technology
    \And
        Przemys\l{}aw Biecek \\
        \texttt{przemyslaw.biecek@pw.edu.pl} \\
        Faculty of Mathematics and Information Science\\
       Warsaw University of Technology,\\
       Faculty of Mathematics, Informatics, and Mechanics\\
       University of Warsaw
       }

\date{}

\maketitle

\begin{abstract}
Benchmarks for the evaluation of model performance play an important role in machine learning. However, there is no established way to describe and create new benchmarks. What is more, the most common benchmarks use performance measures that share several limitations. For example, the difference in performance for two models has no probabilistic interpretation, there is no reference point to indicate whether they represent a~significant improvement, and it makes no sense to compare such differences between data sets.
We introduce a~new meta-score assessment named Elo-based Predictive Power (EPP) that is built on top of other performance measures and allows for interpretable comparisons of models.
The differences in EPP scores have a probabilistic interpretation and can be directly compared between data sets, furthermore, the logistic regression-based design allows for an assessment of ranking fitness based on a deviance statistic.
We prove the mathematical properties of EPP and support them with empirical results of a~large scale benchmark on 30 classification data sets and a real-world benchmark for visual data.
Additionally, we propose a Unified Benchmark Ontology that is used to give a uniform description of benchmarks.
\end{abstract}

\section{Introduction}

The current rapid development of machine learning area has resulted in a considerable increase in the number of new algorithms that need to be compared to the state-of-the-art ones. Along with this emerged the need to establish procedures for the systematic comparisons of algorithms. The current practice is to create benchmarks on predefined sets of tasks, for example GLUE \citep{wang2019glue}, SuperGlue \citep{wang2019superglue}, or VTAB \citep{zhai2020largescale}. Benchmarks are created especially in deep learning because the aim here is to build unified algorithms with understanding beyond the shallow patterns in data, which is why it is crucial to compare algorithms on a~wide variety of tasks from different domains.
Another way to evaluate the progress in algorithms development are biological competitions, such as Critical Assessment of protein Structure Prediction (CASP) \citep{casp}, Critical Assessment of protein Function Annotation algorithms (CAFA) \citep{Zhou2019}, or Critical Assessment of PRediction of Interactions (CAPRI) \citep{capri}. Through this, the performance of algorithms for predicting new structures, properties, or interactions between proteins is regularly compared.
Another popular approach for model comparison is storing and sharing the results of multiple algorithms on multiple data sets on platforms such as Kaggle\footnote{\url{https://www.kaggle.com/}}, Papers With Code\footnote{\url{https://paperswithcode.com/}} or OpenML \citep{OpenML2013}.

These days, it is difficult to imagine a high-quality article with a new algorithm without comparing it with the state-of-the-art methods on at least one of the benchmarks. Despite that, there is no unified description of benchmarks to refer to when describing new ones. What is more, performance measures of models currently used in benchmarks share many limitations, such as the lack of possibility to interpret differences in performance or the impossibility of comparing models between data sets.

Considering the shortcomings of existing benchmarks, the need for new approaches for comparing models and establishing new guidelines is being felt in the machine learning community \citep{MartnezPlumed2021}. This urgency is borne out by the fact that in 2021, the organizers of the Thirty-fifth Conference on Neural Information Processing Systems provided a new track dedicated to data sets and benchmarks\footnote{\url{https://nips.cc/Conferences/2021/CallForDatasetsBenchmarks}}.

In this article, we propose an Elo-based Predictive Power (EPP) Meta-Score, which is a~new way of aggregating model results and which overcomes the most common problems with benchmarks and the performance scores they use. 
The main contributions of this paper are as~follows.
\begin{itemize}
    \item We propose a Unified Benchmark Ontology that allows for the uniform description of different benchmarks.
    \item We identify and demonstrate the limitations of the most common measures of machine learning model performance, such as the lack of interpretation of differences and incomparability between data sets.
    \item In light of the highlighted limitations of the most common measures, we propose a~new Meta-Score named EPP that is built as an aggregation of other measures and enriches them by providing interpretable comparisons of models, even between data sets.
    \item We apply EPP on a~large-scale benchmark from the OpenML repository and Visual Task Adaptation Benchmark. In both cases, we show how the use of EPP enriches the understanding of model performance.  
\end{itemize}

\subsection{Historical overview}
The problem of model assessment is even older than modern statistics. Its origins can be traced to Laplace's work from 1796 on the nebular hypothesis. Since then, the increasing number of applications for models has led to an increase in the number of metrics describing their quality. The various measures of model performance differ in their properties and applications \citep{PowersD, SOKOLOVA2009427}. The most common machine learning frameworks such as scikit-learn \citep{scikit-learn}, TensorFlow \citep{tensorflow2015-whitepaper}, or mlr \citep{mlr} rely on common measures such as accuracy, AUC, Recall, Precision, F1, cross-entropy for classification and MSE, RMSE, MAE for \mbox{regression problems}. 

Beyond measure selection, there is an even more important problem related to~evaluating whether the~differences between its values are significant, or whether they come from noise in~validation data sets.
There have been many approaches to verify whether a new proposed algorithm improves the performance compared to previous state-of-the-art methods. The majority of them have been statistical testing procedures. Janez Dem{\v{s}}ar \citep{janez_2006} reviewed commonly used practices and point out the vast number of problems with them. 
One of the earliest and most widely cited articles in this area is the one by Dietterich \citep{Dietterich:1998:AST:303222.303237}. He~gave a~broad description of the~taxonomy of the different kinds of statistical questions that arise in~machine learning. He also introduced a~new procedure for testing which of two classifiers is more accurate, called 5x2cv t-test and based on 5 iterations of 2-fold cross-validation. In
each replication, two algorithms are trained on each fold
and tested on the other fold. The test statistic is then a modified statistic of a paired t-test, where the standard deviation comes from the cross-validation.  The 5x2cv test was later improved.   Alpaydin \citep{Alpaydin:1999:CTC:1121924.1121932} introduced a robust 5x2cv F test while Bouckaert \citep{Bouckaert:2003:CTL:3041838.3041845} doubted the theoretical degrees of freedom and corrected them due to dependencies between experiments. 
However, the above methods do not fit into the actual trends in machine learning algorithms, where new algorithms are tested against multiple state-of-the-art models and over several data sets. For this purpose, more extensive methods began to be used.
One of them is ANOVA \citep{Salzberg1997}, for example with Friedman's test \citep{10.1007/3-540-45723-2_10, PIZARRO2002155} for comparison of multiple models, however it can only give conclusions about whether there are differences in the performance of models. If there are differences, a post-hoc tests are needed to determine which model has better performance. One of the procedures used for post-hoc analysis is the Nemenyi test, that gives the statistical significance of performance differences for a pair of models. Results of Nemenyi test for many models and may be aggregated as Critical Difference plot that shows groups of models that are not different. However, these results are not transferable to new data sets (not used during testing) because tests results have no absolute meaning. Therefore, once we do a ranking, we cannot extend it without repeating the entire testing procedure.
The first to use non-parametric tests for comparing models on multiple data sets was Hull \citep{hull1994information}.
Brazdil and Soares \citep{10.1007/3-540-45164-1_8} used ranks to compare classification algorithms, yet they do not provide \mbox{statistical tests}.

Dem{\v{s}}ar \citep{janez_2006} analyzed papers from five International Conferences on Machine Learning (1999-2003) that compared at least two classification models. The conference papers included a~wide range of approaches, from naive average accuracy over all data sets, through counting the number of times a~model performed better than the~others, to assessing statistical significance by pairwise t-tests. However, despite multiple hypothesis testing, only a few articles had Bonferroni correction, that is, a method to adjust tests' p-values in case of multiple comparisons.

The conclusion of the analysis was that there is no well-established procedure for comparing algorithms over multiple data sets. Furthermore, there are issues with common measures of model performance, such as uninterpretable differences between two values, or the inability to compare these values between data sets (see Section~\ref{sec:ml_problems}). Therefore, there is an emerging need to develop better solutions for models benchmarking. In this paper, we introduce a method of model comparison that is based on the Elo ranking system used in sports, for example in chess and football.

\subsection{Elo ranking system}
\label{sec:elo}

The rating introduced by Elo \citep{elo2008rating} is a~ranking system used for calculating the relative level of a player's skill.
The difference between the Elo ratings of two players can be transferred into the probabilities of winning when they play against each other. Therefore, the~difference in Elo scores is a~predictor of the match result calculated on the basis of the history of players' matches.
 The scores of players are updated after each match they play, and a new rating is calculated on the basis of two components: the result of a~match and the rating of the opponent. A~player's level is not measured absolutely, although it is inferred from wins, losses, and draws against other players. After each match, the winner gains Elo points. The amount of received points is related to the strength of the opponent. If a~player beats an opponent that has a higher Elo score, the victor  would gain more points than if playing a weaker opponent. Conversely, a~defeated player would lose more points if he lost a~match against a player with a lower rating.

An Elo rating has many variations; one of them, popular in chess, is the algorithm of 400. It states that an average player has a~rating of 1500, and reaching a rating over 2000 means that the player is one of the best. Let us consider two players, $Player_1$ and $Player_2$. $Player_1$ has an expected score of $E_1$, which is their probability of winning plus half of the probability of drawing with $Player_2$, and is expressed as:

 $$E_1 = \frac{1}{1 + 10 ^{\frac{\left(S_1-S2\right)}{400}}},$$ 

where $S_1$ and $S_2$ are ratings of $Player_1$ and $Player_2$.
This formula shows an important property of Elo scores - the possibility to interpret them in terms of the probability of winning.
For example, the difference of 200 rating points means that a more skilled player has a probability of winning $\frac{1}{1+10^{-\frac{200}{400}}} \approx  0.75$.

In addition to probabilistic interpretation, Elo rating has one more advantage. It is not necessary for every player to play against each other to provide a comparison of their skills. In the real world, it would be impossible to stage matches between all chess players, therefore Elo is used to find an approximation of true skill. Of course, the more matches played, the better the approximation; however, not all players need to play against each other.

\section{Unified Benchmark Ontology}
\label{sec:ontology}

\begin{figure}[!hbt]
\centering
    \includegraphics[width=0.8\textwidth]{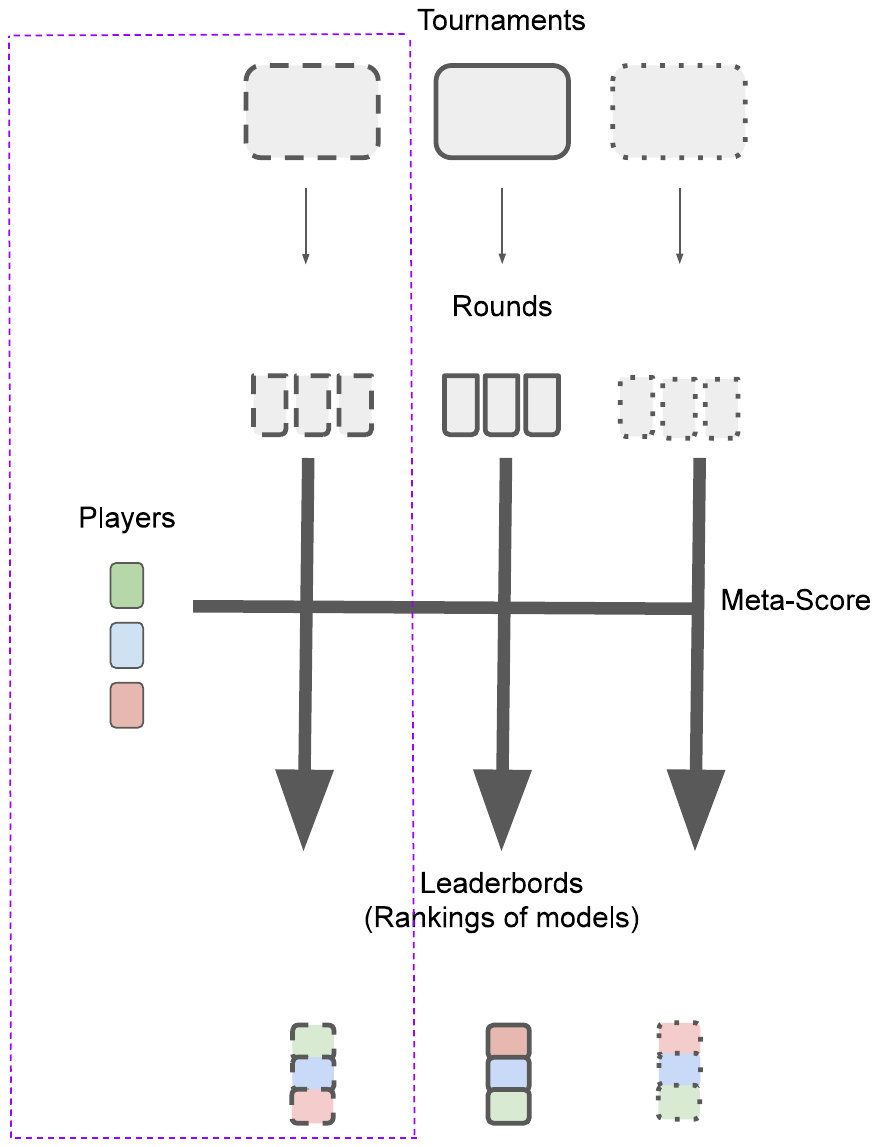}
    \vspace{-0.5cm}
    \caption{A Unified Ontology of ML Benchmarks. The violet dashed rectangle shows a minimal setup for any benchmark.}
    \label{fig:ontology}
\end{figure}

In this section, we introduce the Unified Benchmark Ontology for machine learning that fills the gap for a uniform description of benchmarks.
Figure~\ref{fig:ontology} contains a unified diagram for describing machine learning benchmarks. We use terms associated with sports tournaments, such as \player, \tournament, \round, \leaderboard. The detailed descriptions of all these components are in Table~\ref{tab:components_ontology}. In this article, whenever we refer to the components of a Unified Benchmark Ontology, we will indicate this with capitalization and italics. Each component may be assigned to a different machine learning element. The set of such assignments is a~\scheme. Examples of \schemes are provided in the next subsection.

\begin{table}[!htb]
\centering
\caption{The descriptions of the EPP Benchmark components.}
\label{tab:components_ontology}
\begin{tabular}{llc}
\toprule
\textbf{Component} & \multicolumn{1}{c}{\textbf{Description}} & \textbf{Example} \\ \midrule
\player$_i$ & \begin{tabular}[c]{@{}l@{}}A single i-th participant of the EPP  Benchmark. \end{tabular} & \begin{tabular}[c]{@{}c@{}}Classification \\ model \end{tabular} \\ \hline
\score & \begin{tabular}[c]{@{}l@{}}A one-dimensional measure of a Player's strength. \\ We assume  that the  order relation over Scores \\ is given and monotonic.\end{tabular} & Accuracy \\ \hline
\round$_r$ & \begin{tabular}[c]{@{}l@{}}A single game environment for \players. The \\outcome of a \round$_r$ are score values of \players.  \end{tabular} & \begin{tabular}[c]{@{}c@{}}Cross-validation \\ fold \end{tabular} \\ \hline
\tournament & \begin{tabular}[c]{@{}l@{}}An independently replicated \rounds.\end{tabular} & Data set \\ \hline
\metascore & \begin{tabular}[c]{@{}l@{}}A measure of a \player's  strength aggregated \\ over all \rounds in a \tournament.\end{tabular} & Mean \\ \hline
\leaderboard & \begin{tabular}[c]{@{}l@{}}The ordering of \players  according to their overall \\ strength on all \rounds in a \tournament.\end{tabular}  & \begin{tabular}[c]{@{}c@{}} Mean Accuracy \\ of models over \\  CV folds\end{tabular} \\ \hline
\scheme & \begin{tabular}[c]{@{}l@{}}An assignment of EPP Benchmark components to \\ specific machine learning pieces.\end{tabular} & --- \\ \bottomrule
\end{tabular}
\end{table}

\subsection{Example Schemes}
\label{sec:schemes}

In Table~\ref{tab:schemes}, we present example schemes, i.e., mappings between components of a Unified Ontology Benchmark and machine learning terms.
\scheme Model/CV is one of the most standard benchmarking settings where models are compared on different cross-validation splits. 
\scheme Model/Task covers a situation when models are compared on several data sets where each is assigned to its own performance measure. The pair data set and performance measure is a task. Examples of such benchmarks are SuperGlue and the Visual Task Adaptation Benchmark.
The third example \scheme is Data Set/Model. The aim here is to compare data sets and assess how high performance models can score on it. This can be useful in assessing how simple it is to train a good model on a particular data set.

\begin{table}[!htb]
\centering
\caption{Example \schemes for EPP Benchmark.}
\label{tab:schemes}
\begin{tabular}{llll}
\toprule
\textbf{Component} & \begin{tabular}[c]{@{}l@{}}\textbf{Scheme  Model/CV}\end{tabular} & \begin{tabular}[c]{@{}l@{}}\textbf{Scheme  Model/Task}\end{tabular} & \begin{tabular}[c]{@{}l@{}}\textbf{Scheme Data Set/Model}\end{tabular} \\ \midrule
\player & Model & Model & Data set \\ \hline
\score & \begin{tabular}[c]{@{}l@{}}Performance \\ measure, for \\ example, AUC\end{tabular} & \begin{tabular}[c]{@{}l@{}}Score defined \\ separately for \\ each data set\end{tabular} & \begin{tabular}[c]{@{}l@{}}Performance \\ measure, \\ for example, AUC\end{tabular} \\ \hline
\round & \begin{tabular}[c]{@{}l@{}}Cross-Validation\\ split\end{tabular} & \begin{tabular}[c]{@{}l@{}}Data sets with one \\ train/test split each\end{tabular} & Model \\ \hline
\tournament & Data set & One data set & Set of models \\ \hline
\leaderboard & \begin{tabular}[c]{@{}l@{}}Separate rankings \\ of models for each \\ data set\end{tabular} & \begin{tabular}[c]{@{}l@{}}One ranking of \\ all models\end{tabular} & \begin{tabular}[c]{@{}l@{}}One ranking of \\ all data sets\end{tabular} \\ \bottomrule
\end{tabular}
\end{table}

Depending on the Scheme assumptions, the \scores for a particular \player may be independent (when \rounds are different data sets) or correlated (when \rounds are cross-validation splits) across \rounds. In both cases, we assume that within \round the \players' \scores are comparable and can be ranked against each others's according to a given order relation.

\section{Elo-based Predictive Power (EPP) Benchmark}
\label{sec:epp}

In this Section, we introduce the EPP Benchmark that fits into the nomenclature introduced in Section~\ref{sec:ontology}. In Section~\ref{sec:epp_definition}, we show key concepts of the Elo-based Predictive Power (EPP) score, while in Section~\ref{sec:ml_problems}, we show the common problems with state-of-the-art benchmarking methods and we derive the properties of the EPP score that overcome such issues.

\begin{figure}[!hbt]
\centering
    \includegraphics[width=1\textwidth]{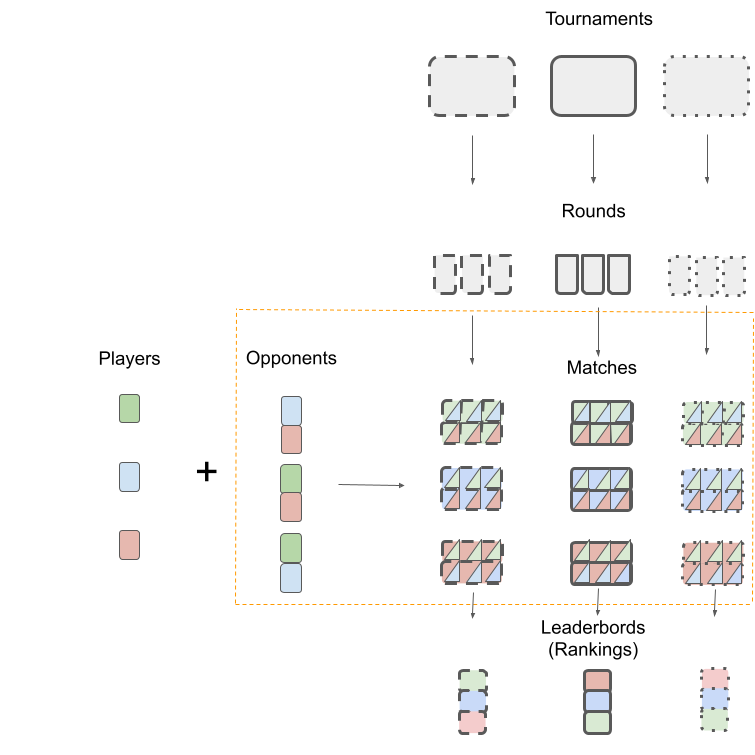}
    \vspace{-0.5cm}
    \caption{A diagram of the EPP Benchmark. The orange dashed lines shows EPP-specific parts of the benchmark and were not included in the unified benchmark.}
    \label{fig:elo_unified}
\end{figure}

Figure~\ref{fig:elo_unified} presents the EPP Benchmark with the nomenclature from the Universal Benchmark Ontology. The thick arrow from Figure~\ref{fig:ontology} is broken down here into additional components, such as \opponents and \matches. The detailed descriptions of components that are specific for EPP Benchmark are in Table~\ref{tab:components}. 

\begin{table}[!htb]
\centering
\caption{The descriptions of the EPP Benchmark components that extends the Universal \mbox{Benchmark Ontology.}}
\label{tab:components}
\begin{tabular}{ll}
\toprule
\textbf{Component} & \multicolumn{1}{c}{\textbf{Description}} \\ \midrule
 \opponent$_{i, j}$ & \begin{tabular}[c]{@{}l@{}} \player$_j$ whose \score values are compared to the \scores values of \\ the \player$_{i}$. 
\end{tabular} \\ \hline
\match$_{i,j,r}$ & \begin{tabular}[c]{@{}l@{}}A single comparison of the \score values of a pair of \players, i.e. \\ \player$_i$ and \opponent$_{i,j}$ in \round$_r$. \end{tabular} \\ \bottomrule
\end{tabular}
\end{table}

\subsection{The Concept of the Elo-based Predictive Power (EPP) Meta-Score}
\label{sec:epp_definition}

Elo-based Predictive Power (EPP) Meta-Score is used for establishing the \players' \leaderboard according to a single \round of the experiment. The \players are ranked according to their \score values in a single \round.
However, the order of \players on the \leaderboard consistent with all \rounds may be impossible to determine. A single order  might not have the property of connectivity with all \rounds, and therefore be nonlinear.  That is why the common procedure is to aggregate, for example as an average, the \scores over \rounds and then obtain the \leaderboard. However, the mean is sensitive to outlier observations, thus models with strongly varying results will distort the aggregated ranking.
In this paper, we introduce an alternative approach that is EPP. The idea is not to aggregate values of \scores, but to compare the relative performances of \players. We ignore the absolute values of \players' \scores and the winner is the one whose \score is better (in terms of given order relation). Every  \round$_r$ consists of \match$_{i,j,r}$ in which \player$_i$ competes with other \player$_j$ (\opponent$_{i,j}$). In consequence, for every pair of \players we get the sequence of win/lose results for every \round and can use these table for calculating the relative EPP Scores of \players' performances. This relativity of \players' performance makes EPP very similar to Elo, in particular in the way that both methods give a probabilistic interpretation of differences in score values.

However, the limitations of Elo ranking used in sports does not apply to the EPP for machine learning benchmarks. In classic Elo ranking, not every \player stands against every other. One hundred \players would have to play $\frac{100 \cdot 99}{2} = 4950$ \matches, which might be impossible for logistical or time reasons. Therefore, it is often hard to use all possible results of \matches. In the case of machine learning models, the cost of calculating EPP \metascores is not as time consuming as human \matches. It is worth noting that a~\match result is a~comparison of \players' performances in one \round and the performance for a~particular \score is the same, regardless of the \opponent. Therefore, for one hundred players we can obtain results of all \matches  calculating only $100$ values (performance of each \player in each \round). It is worth noting that it is not always possible to get \score value for each model in every \round, for example due to the missing data that only some of the models can deal with. However, EPP can still be calculated in the presence of missing \players' \scores.

In classic Elo ranking, the scores are updated after consecutive matches, therefore there is a~natural order of updates. As the Elo points by which the winning player's score increases depends on his Elo and the opponent's Elo, the order in which the matches are played may affect the Elo's final score.
However, it should be noted that the need to sequentially calculate Elo is due to the aforementioned weakness, which is the inability to play matches between all players at once.
We propose an EPP model scoring method that does not require sequential calculation of match results and preserves the desired Elo properties, i.e. the possibility of interpretation on an interval scale.
 
It is worth noting that Elo \citep{elo2008rating} proposed a solution to the problem of how to measure the skill of all players with only partial information about the outcome of matches. The EPP score applied to machine learning models take into consideration all results (we have a measure of performance for all models) and therefore is a~direct way to calculate the values approximated with Elo. For that reason, the order of model comparisons is irrelevant for~EPP.

\subsection{Definiton of the EPP score}
 
Now, we formally define the EPP meta-score in terminology of the unified EPP benchmark. Let $\mathcal{M} = \{ M_1, M_2, ..., M_m \}$ be a~set of $m$ \players. For a~selected single \tournament $T_t$, we specify a~set of $r$ \rounds $\mathcal{R} = \{  R_1, R_2, ..., R_r \}$ and \score.

Let denote  the result of a single \match in a \round $R_k$ between \players $M_i$ and $M_j$ as

\begin{align*}
    y_{i,j}^{k} = \begin{cases}
    1 ,& \text{where \player~} M_i \text{ has better \score value than \player~} M_j, \text{in \round~} R_k \\
    0.5, & \text{where \player } M_i \text{ has the same \score value as \player  } M_j,  \text{in \round~} R_k\\
    0, & \text{otherwise,}
    \end{cases}
\end{align*}

and $\sum_{k=1}^r y_{i,j}^k$ is the number of wins the \player $M_i$ over the \player $M_j$ in all \rounds. The \scores are usually continuous, so the probability of a tie is near 0.
Therefore, the empirical probability of winning in a random \round is equal 

\begin{align*}
    p_{i,j} = \frac{\sum_{k=1}^r y_{i,j}^k}{r}.
\end{align*}

\begin{definition}
 The odds(i,j) are odds that \player $M_i$ has a better \score than \player $M_j$, and are expressed~as
 
$$ odds(i,j)~=~\frac{ p_{i,j} }{1 - p_{i,j} },$$
 
 where $p_{i,j}$ is the probability that \player $M_i$ has a better \score than \player $M_j$ in a~random~\round $R$.
\end{definition}

\begin{definition}[]
\label{def:epp}
The $\beta_{M_i}$ and $\beta_{M_j}$ are EPP \metascores for \players $M_i, M_j \in \mathcal{M}$ respectively if they satisfy the following property

$$\log{\frac{ p_{i,j} }{1 - p_{i,j} }} = \beta_{M_i} - \beta_{M_j} ,$$

where $p_{i,j}$ can be estimated $\hat{p}_{i,j}$ in two exploratory variables logistic regression of the form
 
$$\log{\frac{ \hat{p}_{i,j} }{1 - \hat{p}_{i,j} }} = \hat{\beta}_{M_i} x_{M_i} + \hat{\beta}_{M_j} x_{M_j}, \hspace{2em}
where \hspace{0.5em}
  x_{M_i} = 1 \text{ and } x_{M_j} = -1,$$ 

where $\hat{\beta}_{M_i}$ and $\hat{\beta}_{M_j}$ are estimated EPP \metascores. For brevity, in the following sections, we refer to them simply as EPP \metascores.
\end{definition}

\begin{definition}
 The EPP \metascore \leaderboard for \tournament $T$ is the set of EPP \metascore values for the set of $m$ \players $\mathcal{M} = \{ M_1, M_2, ..., M_m \}$ is
 
 \begin{align*}
     L_{\mathcal{M}}^T = \left(\hat{\beta}_{M_1}, \hdots, \hat{\beta}_{M_m}\right).
 \end{align*}

\end{definition}

The properties of EPP \metascore are in the next subsection.

\subsection{EPP handles the problems with common ML performance measures}
\label{sec:ml_problems}

In this Section, we identify problems with the most common performance measures in ML benchmarks and we show that the EPP \metascore handles these issues. The attributes of EPP \metascore may be described with three aspects:
\begin{itemize}
    \item EPP is a meta-approach based on values of other performance measures. EPP broadens the possibilities of comparing \players because of its unique properties introduced in Sections~ \ref{q:1} - \ref{q:3};
    \item EPP is an alternative approach to aggregating results, such as mean scores for repeated measurements - \rounds. EPP gives the statistical possibility to assess the stability of \scores (see Sections~ \ref{q:4}  and \ref{q:5});
    \item  Unlike the methods used so far, EPP gives the possibility to compare Benchmarks. It allows the assessment of the quality of \leaderboards across \tournaments. (see Section  \ref{q:6}).
\end{itemize}

The following Sections are constructed as follows: first we discuss a~problem with a~real-life benchmark using terminology from the field of machine learning. In the second part of each Section, we discuss at a~general level and describe theoretical properties of the EPP that addresses the problem at hand. We describe such properties in terms of the EPP Benchmark. This distinction helps to better separate the examples from the theoretical part.

\subsubsection{There is no interpretation of differences in performance}
\label{q:1}

\begin{table}[!htb]
\small
\caption{Springleaf Marketing Response Kaggle Competition, \url{https://www.kaggle.com/c/springleaf-marketing-response}}
\label{tab:kaggle_springleaf}
\centering
\begin{tabular}{ll}
\toprule
Team Name                              & AUC   \\ 
\midrule
Asian Ensemble                         & 0.80925 \\
ARG eMMSamble                               & 0.80907 \\
.baGGaj. & 0.80899 \\
\bottomrule
\end{tabular}
\end{table}

In Table~\ref{tab:kaggle_springleaf}, we show an example of Kaggle ranking. 
The difference between AUC value of the~first and AUC value of the second model equals $0.00018$. This absolute~difference gives us no additional information.
The AUC is useful for ordering models, but its differences have no interpretation, it does not provide any quantitative comparison of models' performances.  There is no single accepted way to compare the~power of enhancement of performance measures. Some say we should equate absolute differences regardless of the absolute values of the score, while others  would suggest analyzing relative improvement. Both ways may lead to opposite conclusions, depending on the absolute value of a performance measure.

This ranking fits to the EPP Benchmark \scheme Model/Task with just one \round. EPP score provides the direct interpretation in terms of probability. 
\begin{property}
\label{property:q:1}
The difference of EPP scores for \players $M_i$ and $M_j$ is the logit of the probability that $M_i$ achieves better performance than $M_j$. 

Indeed, from Definition~\ref{def:epp} we have that 
\begin{equation}
\label{eq:log_reg_short}
logit\left(\hat{p}_{i,j}\right) = \log \left(\widehat{odds}(i,j)\right) = \hat{\beta}_{M_i} - \hat{\beta}_{M_j}.
\end{equation}

\end{property}

After reformulating Equation~\ref{eq:log_reg_short} we achieve direct formula for probability that \player $M_i$ achieves better performance than \player $M_j$:

$$
\hat{p}_{i,j} = invlogit\left(\hat{\beta}_{M_i} - \hat{\beta}_{M_j}\right) =  \frac{\exp\left({\hat{\beta}_{M_i} - \hat{\beta}_{M_j}}\right)}{1 + \exp\left({\hat{\beta}_{M_i} - \hat{\beta}_{M_j}}\right)}.
$$

\subsubsection{There is no procedure for assessing the significance of the difference in performances}
\label{q:2}

In Table~\ref{tab:kaggle_fraud}, there are results of an IEEE-CIS  Fraud  Detection  Kaggle  Competition. 
The AUC values of all models in Table~\ref{tab:kaggle_fraud} differ in the third decimal place. There is no reference point to indicate whether this difference represents a~significant improvement in prediction or not. Significance in the statistical sense means these differences are not on the noise level. 

\begin{table}[!htb]
\caption{IEEE-CIS Fraud Detection Kaggle Competition, \url{https://www.kaggle.com/c/ieee-fraud-detection}}
\label{tab:kaggle_fraud}
\centering
\begin{tabular}{ll}
\toprule
Team Name                      & AUC  \\ 
\midrule

AlKo                          & 0.968137 \\
FraudSquad                    & 0.967722 \\
Young for you                    & 0.967637 \\
\bottomrule
\end{tabular}

\end{table}

The scoring of models in this example are deployed in a \scheme Model/Task with just one \tournament and one \round. A similar situation with minor differences appears in many state-of-the-art benchmarks. When a \player gains improvement by a decimal place it would be desired to distinguish between real improvement and apparent improvement due to the noise coming from different \round setting e.g. splitting to train and test data.
Currently, there are not many formal methods to assess the significance of differences. One way is to use a Kruskal-Wallis test for the equality of medians. But results from statistical tests are not transitive.  We can compare two \players, but we would not get an overall \leaderboard for all of them.

EPP score allows for the assessment of the significance of score value, which gives an intuition whether the difference in performance is a~noise or not.
\begin{property}
\label{prop:logreg}
The values of the EPP \metascores are coefficients of logistic regression model with intercept $\beta_{0} = 0$.\\
The Equation~\ref{eq:log_reg_short} can be generalized to 
\begin{equation}
\label{eq:log_reg}
logit(\hat{p}_{i,j}) = \hat{\beta}_{M_1} x_{M_1} + \hat{\beta}_{M_2} x_{M_2} + ... + \hat{\beta}_{M_k} x_{M_n}, \hspace{2em}
where \hspace{0.5em}
  x_{M_a} = 1_{a=i} - 1_{a=j}.
\end{equation}

$\hat{\beta}_{M_i}$ is the estimation of unknown $\beta_{M_i}$ coefficients from the multiple exploratory variables logistic regression where $x_{M_a}$ indicates if the \player is compared.
\end{property}

Because of calculating values of EPP \metascore from logistic regression, a~logit of probabilities gives an additional benefit in the form of gaining a~significance of EPP scores. This is an advantage over raw empirical probabilities. 

\begin{property}
\label{prop:signf_test}
The statistical significance of the difference between EPP for two \players $M_i$ and $M_j$ may be tested as the null hypothesis that 

$$\hat{\beta}_{M_i} = \hat{\beta}_{M_j}.$$ 

If \round performances are independent and sample size is sufficiently, this hypothesis may be tested with Wald test or Likelihood ratio test.
\end{property}

However, even when assumptions about independence of splits are violated and observations appear in different bootstrap samples, one can rely on tests results as they are robust. Another way is to use approximately unbiased bootstrap resampling  \citep{Shimodaira2002,shimodaira2004,10.1093}.

\subsubsection{There is no way to compare \scores between \tournaments}
\label{q:3}

In Tables~\ref{tab:kaggle_springleaf}~and~\ref{tab:kaggle_fraud} differences between second and third best models for each data set are around $0.00008$. The question arises as to whether these differences are comparable between data sets. 
Does $0.00008$ on Springleaf Marketing data mean the same increase of model quality on IEEE-CIS Fraud~data? 

There are at least three points of view. One is that the gaps are almost the same for both data sets, because the differences in AUC values are almost the same.
The second is that the gap in the IEEE-CIS Fraud Competition is larger as the AUC value is close to 1. Relative improvement for Fraud detection 
$\left( \frac{0.967722 - 0.967637}{1- 0.967722} \approx 0.0026\right)$ 
is larger than relative improvement for Springleaf Marketing
$\left( \frac{0.80907 - 0.80899}{1 - 0.80907} \approx  0.0004 \right)$.
The third point of view is that the gap between first and second place for Springleaf ($0.00018$) is smaller than the same difference for IEEE-CIS Fraud detection ($0.000415$). Therefore, the relative gain from the difference between second and third place for Springleaf is higher.

From the definition of EPP score, the probability of winning against an average \player (equivalent to an intercept $\hat{\beta}_0$) has the same meaning, regardless of the \tournament. The EPP scores are absolute values with a mean equals to zero. Therefore, comparison of EPP values between \tournaments is possible by comparing a~probability of winning against an average~\player.

\begin{property}
\label{property:q:3}
Probability that \player $M_i$ would win against an average \player $M_{avg}$ is 

$$    \hat{p}_{i, avg}  = \frac{\exp\left({\hat{\beta}_{M_i}}\right)}{1 + \exp\left({\hat{\beta}_{M_i}}\right)},$$

from the Property~\ref{prop:logreg} we have that intercept  $\hat{\beta}_0 = 0$. In the logistic regression, intercept relates to the mean, therefore $\hat{\beta}_{M_{avg}} = 0$ 
and 

$$ \hat{p}_{i, avg} = invlogit\left(\hat{\beta}_{M_i} - \hat{\beta}_{avg}\right) =\frac{\exp\left({\hat{\beta}_{M_i} - \hat{\beta}_{M_{avg}}}\right)}{1 + \exp\left({\hat{\beta}_{M_i} - \hat{\beta}_{avg}}\right)} = \frac{\exp\left({\hat{\beta}_{M_i}}\right)}{1 + \exp\left({\hat{\beta}_{M_i}}\right)}.
$$
\end{property}

\subsubsection{Mean aggregation may be misleading: the Variance for \round}
\label{q:4}

\begin{figure}[!hbt]
\centering
    \includegraphics[width=0.8\textwidth]{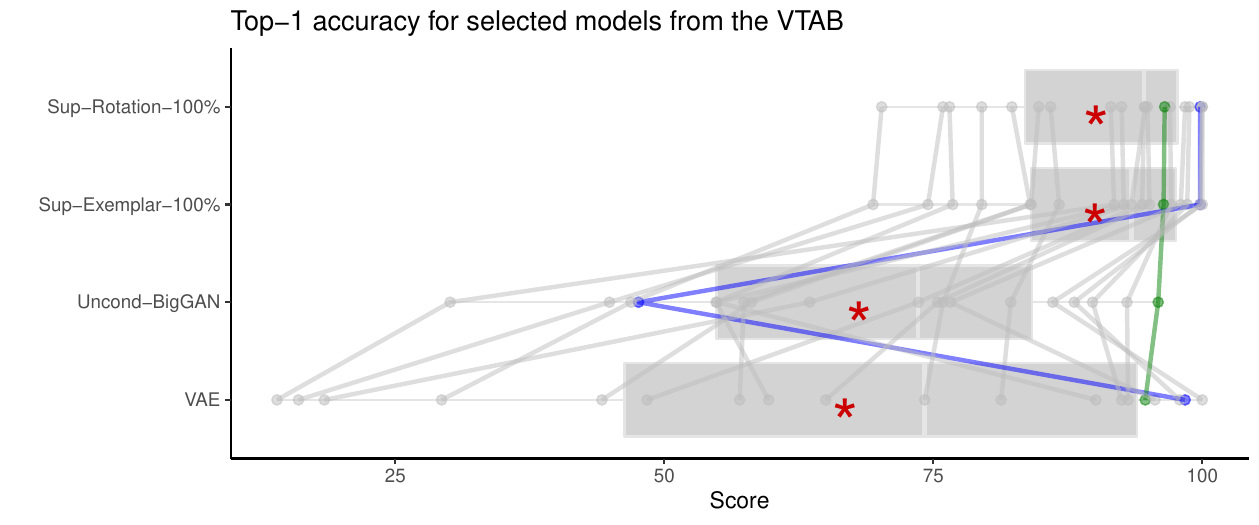}
    \caption{Boxplots of scores for four selected models from the Visual Task Adaptation Benchmark. Red stars correspond to mean scores across all VTAB tasks. Dots indicate scores in separate VTAB tasks. Thin lines connect points corresponding to the same task. Green and Blue lines are highlighted for example purposes in Section~\ref{q:5}.}
    \label{fig:boxplots_vtab}
\end{figure}

In Figure~\ref{fig:boxplots_vtab}, we show four selected models from the Visual Task Adaptation Benchmark (VTAB) \citep{zhai2020largescale}.  Every small point indicates the top-1 accuracy for one model on one of 19 specified data sets. The scores corresponding to the same data set are connected with the thin lines.

We analyse the results of these models within pairs. The first pair is the Sup-Rotation-100$\%$ model and Sup-Exemplar-100\% model. The second pair is  Uncond-BigGAN model and VAE model. Averaged top-1 accuracy across data sets are close to to models in these pairs. The Sup-Rotation-100\% and Sup-Exemplar-100\% have evidently higher predictive power than Uncond-BigGAN and VAE models for \mbox{most tasks}.

Top-1 accuracy scores of Sup-Rotation-100\% and Sup-Exemplar-100\% models are very close to each other within specified data sets. This is represented as parallel lines connecting dots for two models in the top part of the plot. 
In the second pair, the relationship between Uncond-BigGAN and VAE is more ambiguous. Comparing just averages across data sets show that the Uncond-BigGAN model is comparable to VAE.  
However, when considering the green path in Figure~\ref{fig:boxplots_vtab}, all four models have similar performance.
On the other hand, when considering the blue path, VAE significantly defeats Uncond-BigGAN and performs comparably to Sup-Rotation-100\% and Sup-Exemplar-100\%. The probable reason for these two scenarios is the different tunability of data sets, which exhibits as diverse variance of top-1 accuracy for different models.


The VTAB ranking exemplifies the \scheme Model/Task of  EPP Benchmark  with just 19 \rounds determined with data sets.
Averaging the \score values neglects the information about the distribution of \score values within \rounds. This is especially relevant in cases where we compare \scores of different definition and range of values for a~sequence of \rounds. 
EPP also ignores the dispersion of \score values for a~specified \round but this simplification comes, by design, from the definition of EPP computing.

\begin{property}
The EPP score is an aggregate over all rounds.\\
By fitting the logistic regression model from Equation~\ref{eq:log_reg} as dependent variables of observations, we use the results of \matches - whether one \player beats another.
\end{property}

\subsubsection{The mean aggregation may be misleading: the variance for \player}
\label{q:5}

As we see from the example of VTAB rankings in Figure~\ref{fig:boxplots_vtab}, variation of VAE scores are higher than in the case of Uncond-BigGAN. Averaging the \score values  neglects the information about the variance of \score values within \player results. Sometimes the standard deviation of \scores is used as description of stability of results but when \scores values come from various distribution depending the \round this summary is incorrect.

EPP benchmark from the definition  unifies the definition of all \matches in the early phase and EPP \metascore  is computed on the base of win/lose results coming from binomial distribution. Because of that, there is the procedure of assessing the stability of EPP \metascores values.

\begin{property}
The $(1-\alpha)\%$ confidence interval of  EPP for \player  $M_i$  is equal

$$CI_{\hat{\beta}_{M_i}}^\alpha  = \left( \hat{\beta}_{M_i} -  z_{\alpha/2} SE_{\hat{\beta}_{Mi}}, \hat{\beta}_{M_i} +  z_{\alpha/2} SE_{\hat{\beta}_{Mi}} \right),$$ 

\noindent where $z_{\alpha/2} $ is the percentile of standard normal distribution.

If \round performances are independent and sample size is sufficient, the standard deviation of EPP score $\left(SE_{\hat{\beta}_{Mi}} \right)$ may be computed with maximum likelihood estimation method. Otherwise, the non-parametric bootstrap may be applied.
\end{property}

\subsubsection{You cannot assess the quality of a \leaderboard}
\label{q:6}

In the VTAB example in Figure~\ref{fig:boxplots_vtab}, mean aggregation summarises the model performance with a~single value.  In Figure~\ref{fig:boxplots_vtab} we see that the residual values between \score value corresponding to a single task and mean value for
Sup-Exemplar-100\% model are much smaller than for Uncond-BigGAN or VAE.  Averaging the  top-1 accuracy values does not provide any statistics on how compatible the aggregated ranking is with rankings related to a single task, in other words how much information is lost in aggregation.

Assessing the quality of aggregation is crucial because the main objective is to create a \leaderboard emulating the relative performance power of models. If aggregated ranking is consistent with ranking for single tasks, it is reliable and this aggregation may be considered representative. Otherwise, the aggregated ranking is not an appropriate approach and a different aggregation measure or aggregation in subgroups of tasks should be considered. So far, aggregation methods which have been  used to summarise many rankings do not give any  measure reflecting the quality of aggregation.

EPP \metascore is computed on the basis of logistic regression models. From the definition of the generalized linear model, there are tools to assess the quality of the aggregation procedure. We may compare the estimated probability $\hat{p}_{i,j}$ with the true values $p_{i,j}$ using likelihood function.

\begin{property}
The deviance of logistic regression for EPP \metascore is a measure of goodness-of-fit of a~\leaderboard $L$ for \players $\mathcal{M}$ and \tournament $T$.
\begin{align*}
 D\left(\mathbf{p}_{\mathcal{M}}^T,\mathbf{\hat{p}}_{\mathcal{M}}^T\right)=2\left( \log \left(\mathcal{L}(\mathbf{p}_{\mathcal{M}}^T, \mathbf{p}_{\mathcal{M}}^T) \right) -\log \left(\mathcal{L}(\mathbf{p}_{\mathcal{M}}^T, \mathbf{\hat{p}}_{\mathcal{M}}^T )\right) \right) ,
 \end{align*}
where $\mathbf{p}_{\mathcal{M}}^T = (p_{i,j})$ is a vector consisting of the actual empirical probability  of winning for every pair of \players $M_i$ and $M_j$ in \tournament $T$ and 
$\mathbf{\hat{p}}_{\mathcal{M}}^T = (\hat{p}_{i,j})$ is a vector consisting of the predicted value of the probability of winning, respectively. $\mathcal{L}\left(\mathbf{p}_{\mathcal{M}}^T, \mathbf{p}_{\mathcal{M}}^T\right)$ stands for  the logistic regression likelihood function for a~saturated model that provides a~separate parameter for each observation and is the best fitted model and 
$\mathcal{L}\left(\mathbf{p}_{\mathcal{M}}^T, \mathbf{\hat{p}}_{\mathcal{M}}^T \right)$ is  the logistic regression likelihood function for a~considered model.
\vspace{0.5cm}

In addition, as the number of rounds $R \to \infty$  deviance  converges in distribution to chi-square distribution $ D\left(\mathbf{p}_{\mathcal{M}}^T,\mathbf{\hat{p}}_{\mathcal{M}}^T\right) \stackrel{\mathcal{D}}{\longrightarrow}\chi^2_{m(m-2)}$, where $m$ is the cardinality of \players set~$\mathcal{M}$.

\end{property}

If the deviance of the \leaderboard is low, the EPP \metascores and respective estimated probability $\hat{p}_{i,j}$ are close to empirical probability $p_{i,j}$ and this \leaderboard is more reliable than~a \leaderboard with high deviance. In general, we cannot provide the absolute threshold that indicates whether the deviance statistic is low and \leaderboard is reliable. Yet, note that asymptotic distribution is chi-square and the number of degrees of freedom depends only on the number of \players, therefore for the same set of \players we can compare the relative quality of \leaderboards in \tournaments.

\begin{property}
\label{prop:deviance_comp}
Given the set of $\mathcal{M} = \{ M_1, M_2, ..., M_m \}$ \players and two \tournaments $T_1$~and~$T_2$, the quality of EPP \metascore \leaderboards $L_{\mathcal{M}}^{T_1} $ and $L_{\mathcal{M}}^{T_2} $ can be compared using the deviance statistics. If 

\begin{align*}
    D\left(\mathbf{p}_{\mathcal{M}}^{T_1},\mathbf{\hat{p}}_{\mathcal{M}}^{T_1}\right) \leq     D\left(\mathbf{p}_{\mathcal{M}}^{T_2},\mathbf{\hat{p}}_{\mathcal{M}}^{T_2}\right),
\end{align*}

the \leaderboard $L_{\mathcal{M}}^{T_1}$ for \tournament $T_1$ is better fitted to actual probabilities than the \leaderboard $L_{\mathcal{M}}^{T_2}$ for \tournament $T_2$.

\end{property}

The number of degrees of freedom of deviance statistic is order of $m^2$, therefore deviance statistics can take very large values if the number of \players $m$ is in the order of $100$ or higher. From $\chi^2$ distribution properties, the skewness of this asymptotic distribution decreases with the number of \players $m$. If the number of \players is sufficiently high, $D\left(\mathbf{p}_{\mathcal{M}}^{T_1},\mathbf{\hat{p}}_{\mathcal{M}}^{T_1}\right)$ converges in distribution to  normal distribution  $\mathcal{N} (m(m-2), 2m(m-2))$. The deviance statistics can be scaled and shifted depending on the number of \players.

\begin{property}
\label{prop:deviance_stand}
Standardized deviance statistics for set of \players $\mathcal{M}_i$ and \tournament $T_j$ is

\begin{align*}
 \widetilde{D}\left(\mathbf{p}_{\mathcal{M}_i}^{T_j},\mathbf{\hat{p}}_{\mathcal{M}_i}^{T_j}\right)=  \frac{D\left(\mathbf{p}_{\mathcal{M}_i}^{T_j},\mathbf{\hat{p}}_{\mathcal{M}_i}^{T_j}\right) - m(m-2)}{\sqrt{2m(m-2)}},
 \end{align*}
 
\noindent where $m$ is the cardinality of \players set~$\mathcal{M}_i$.

\vspace{0.5cm}

If the number of rounds $R \to \infty$ deviance and  $m$ is sufficiently large,  $  \widetilde{D}\left(\mathbf{p}_{\mathcal{M}_i}^{T_j},\mathbf{\hat{p}}_{\mathcal{M}_i}^{T_j}\right) \stackrel{\mathcal{D}}{\longrightarrow} \mathcal{N}(0,1)$.

\end{property}

It is worth emphasizing that the transformation of deviance statistics is done by constants, depending only on the number of \players. We do not need to estimate additional coefficients.

Due to the approximation with a standard normal distribution, we can compare the loss between estimated probabilities and the observed values in  different \schemes of \tournaments, but not necessarily for the same set of \players.

\begin{property}
\label{prop:deviance_stand_comp}
Given the set of $\mathcal{M}_1 = \{ M_1^1, M_2^1, ..., M_m^1 \}$ \players in \tournament $T_1$ and~the set of $\mathcal{M}_2 = \{ M_1^2, M_2^2, ..., M_n^2 \}$ \players in \tournament $T_2$, the quality of EPP \metascore \leaderboards $L_{\mathcal{M}_1}^{T_1} $ and $L_{\mathcal{M}_2}^{T_2} $ can be compared using the deviance statistics. If 

\begin{align*}
    \widetilde{D}  \left( \mathbf{p}_{\mathcal{M}_1}^{T_1},\mathbf{\hat{p}}_{\mathcal{M}_1}^{T_1} \right) \leq     \widetilde{D}\left(\mathbf{p}_{\mathcal{M}_2}^{T_2},\mathbf{\hat{p}}_{\mathcal{M}_2}^{T_2}\right),
\end{align*}

\noindent the \leaderboard $L_{\mathcal{M}_1}^{T_1}$ for \tournament $T_1$ is better fitted to actual probabilities than \leaderboard $L_{\mathcal{M}_2}^{T_2}$ for \tournament $T_2$. Which means that \leaderboard  $L_{\mathcal{M}_1}^{T_1}$ captures the relationship between models' performance better than \leaderboard $L_{\mathcal{M}_2}^{T_2}$.   \end{property}

An alternative approach to comparing deviance statistics is the comparison of p-values corresponding to chi-square distribution  (Property~\ref{prop:deviance_comp}) or for standard normal distribution (Property~\ref{prop:deviance_stand_comp}) for deviance statistics. The higher p-value corresponding to deviance statistics indicates that the leaderboard fits more accurately. However, with the increase of the number of \players, we may observe the discretization of p-values related to deviance for tournaments.

 \section{Real data examples}
 In this section, we show that EPP \metascore improves the existing real data benchmarks for tabular data as well as Computer Vision and Natural Language Processing problems. 
 
Firstly, we regard the OpenML100 benchmark for table data \citep{bischl2017openml}, which fits the Model/CV scheme from Section~\ref{sec:schemes}. Every model (\player) is tested on a different train/test split of cross-validation (\round). The rankings are created per data set (\tournament).

There is a different perspective  in VTAB and SuperGlue benchmarks. Similarly to the OpenML, the compared items are different neural architectures - \players. But the \rounds are determined by different independent tasks.

\subsection{Computing EPP on the OpenML100 benchmark}
\label{sec:experiments}

Now, we demonstrate the advantages of EPP \metascores in \leaderboards on a MementoML \citep{kretowicz2020mementoml} database, that is a~large-scale benchmark on 30 binary classification data sets from the OpenML. 
 We selected 5 machine learning algorithms: gradient boosting machines (\texttt{gbm}), a generalized linear model with regularization (\texttt{glmnet}), k-nearest neighbours (\texttt{kknn}), and two implementations of random forest (\texttt{RF} and \texttt{ranger}). Each algorithm was trained with 400 different, randomly chosen hyperparameter configurations. For each data set, models were tested on 20 random train/test splits with AUC as a performance measure. This gave us an overall number of AUC values equal to $30 \cdot 20 \cdot 5 \cdot 400 = 1 200 000$.

On the computed AUC \scores, we calculated the EPP \metascore introduced in Section~\ref{sec:epp} and Figure~\ref{fig:elo_unified}.~As~a~single \round, we consider a train/test split. A \match is a comparison of performances of two models with specified hyperparameters on the same data set, yet not necessarily on the same train/test split.  
As a~result, we have obtained EPP values for each data-model-hyperparameter combination, which gave us $30 \cdot 5 \cdot 400 = 60 000$ values of EPP \metascores.

 \begin{figure}[]
\centering
    \includegraphics[width=\textwidth]{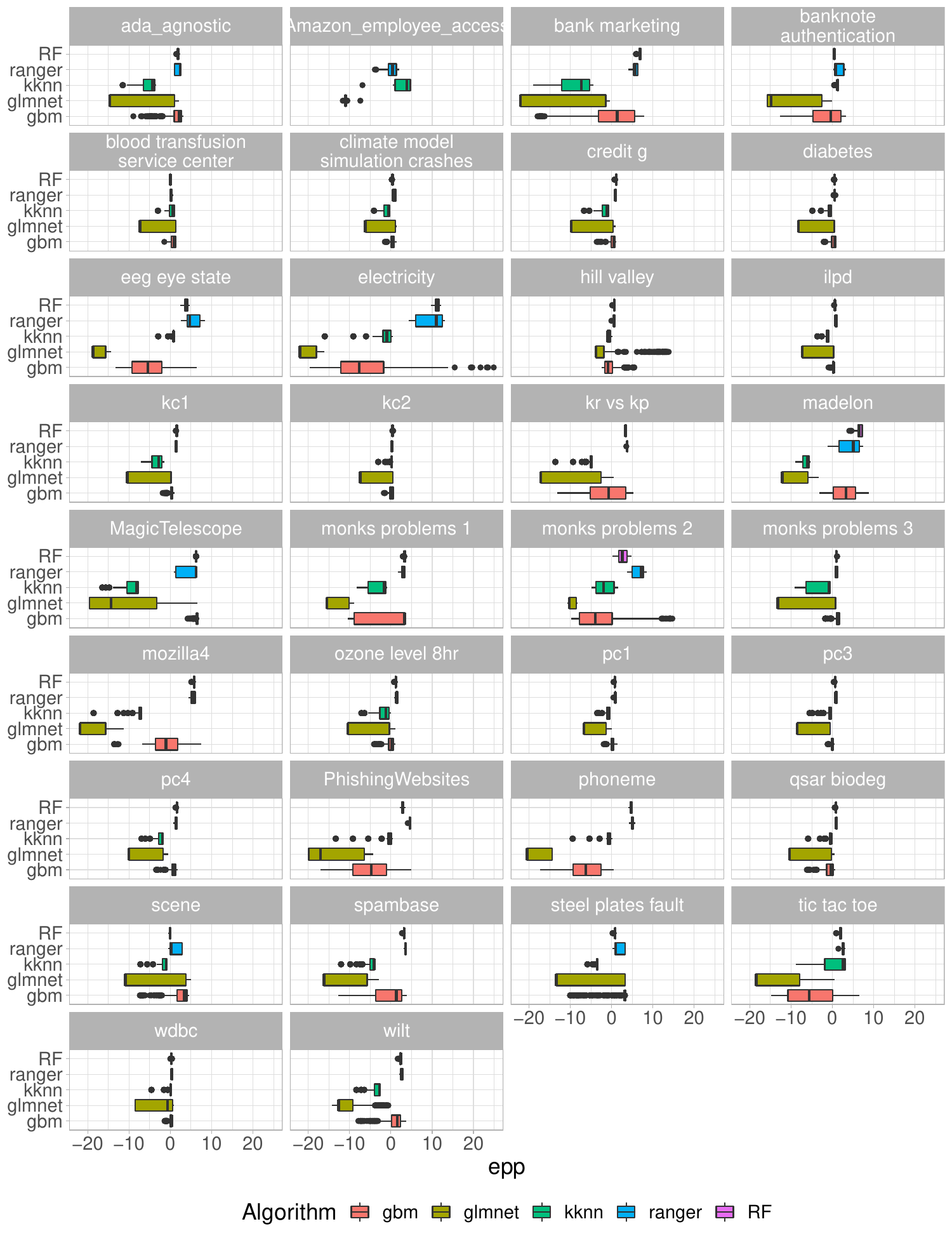}
    \caption{Boxplots of EPP scores for different algorithms across data sets. Each boxplot aggregates EPP scores of all models trained on all data sets.}
    \label{fig:grid_elo_boxplots_part}
\end{figure}

The performance of models is highly variable due to the data set, which can be seen in Figure~\ref{fig:grid_elo_boxplots_part}, where the distributions of EPP \metascore values across models and data sets are shown. The longer boxplot means greater potential for model tuning; for example, we can see that tree-based models (\texttt{gbm} , \texttt{RF}, and \texttt{ranger}) perform better on data set \textit{madelon} than the other two models. Also, all of the EPP values for random forest are positive, which means that, generally, the performance of random forest is above the average.
Due to the independent sampling of hyperparameters and an excessive  $L_1$ penalty in regularization, a~part of the \texttt{glmnet} models achieved AUC score equal to 0.5 or less.  The models with AUC=$0.5$ always lose against other models, which causes a~huge range of values of EPP scores for \texttt{glmnet}.

In Table~\ref{tab:example_epp_results}, we show AUC and EPP values for the four selected models for the \textit{ada\_agnostic} data set from experiments described earlier. To recall, in Section~\ref{q:1}, in the example of Kaggle ranking, we postulate that AUC score does not provide a probability interpretation. EPP addresses this issue, so we can assess the probability of one model winning against another model according to Property~\ref{property:q:1}. 
The descending order according to the averaged AUC is different from EPP ranking. The lowest EPP value has the \texttt{kknn} model, even though the lowest averaged AUC corresponds to \texttt{glmnet}.
The difference between AUC of the first and AUC of the second model equals $0.001497$,  the difference between AUC of the third model and the fourth score equals $0.00395$.  Due to EPP, we can estimate the probability that \texttt{gbm} beats \texttt{ranger} with $logit{(1.27  - 1.08)} \approx 0.55$ probability. In the second pair of \texttt{glmnet} and \texttt{kknn} models, despite the close to averaged  AUC, there is a $0.83$ likelihood that \texttt{glmnet} will defeat the \texttt{kknn} model. These dissimilarities are not emphasized by AUC score, since the averaged crossvalidation scores miss the variability of metrics.

\begin{table}[ht]

 \small
 \caption{EPP of  selected models for \textit{ada\_agnostic} data set. AUC values \mbox{are averaged.}. The numbers of models are IDs from the MementoML benchmark.}
\label{tab:example_epp_results}
\centering
 \begin{tabular}{lcc}
\toprule
 Model &  AUC        & EPP       \\ 
  \midrule
 \texttt{gbm1305} & 0.890     & 1.27   \\ 
 \texttt{ranger1088} & 0.888  & 1.08   \\ 
 \texttt{kknn1396} & 0.816    & -7.52 \\ 
 \texttt{glmnet1242} & 0.812  & -5.91  \\ 
\bottomrule
\end{tabular}
\end{table}

 \begin{table}[ht]

\small
\caption{The best models in algorithm class for \textit{mozilla4} data set. AUC values \mbox{are averaged.} The numbers of models are IDs from the MementoML benchmark.}
\label{tab:comp_2_datasets_df_1_epp}
\centering
\begin{tabular}{lcc}
\toprule
 Model & AUC & EPP \\ 
  \midrule
 \texttt{gbm1184}          & 0.986 & 7.49 \\ 
 \texttt{ranger1106}       & 0.984 & 6.25 \\ 
 \texttt{RF1106}           & 0.984 & 6.22 \\ 
 \texttt{kknn1016}         & 0.942 & -6.78 \\ 
 \texttt{glmnet1011}       & 0.922 & -11.24 \\ 
\bottomrule
\end{tabular}
\end{table}

\begin{table}[ht]
\small
\caption{The best models in algorithm class for \textit{credit-g} data set. AUC values \mbox{are averaged.} The numbers of models are IDs from the MementoML benchmark.}
\label{tab:comp_2_datasets_df_2_epp}
\centering
\begin{tabular}{lcc}
\toprule
 Model & AUC & EPP \\ 
\midrule
  \texttt{RF1155}        & 0.809 & 1.29 \\ 
  \texttt{ranger1212}          & 0.807 & 1.16 \\ 
  \texttt{gbm1136}           & 0.807 & 1.16 \\ 
  \texttt{glmnet1379}         & 0.802 & 0.97 \\ 
  \texttt{kknn1038}          & 0.769 & -0.54 \\ 
\bottomrule
\small
\end{tabular}
\end{table}

With respect to Property~\ref{property:q:3}, EPP \metascore enables the  analysis of performances between datatsets. Because of the lack of interpretation of AUC differences, comparison between model scores may be made in various ways, as described in detail in Section~\ref{q:3}.  Table~\ref{tab:comp_2_datasets_df_1_epp} and Table~\ref{tab:comp_2_datasets_df_2_epp} present rankings for best-in-class models for two data sets from our experiment, \textit{mozzilla4} and \textit{credit-g}. Even though absolute differences of AUC between the first and second model in each ranking are around $0.002$, the rankings have different levels of AUC scores (approximately $0.98$ and $0.81$ for \textit{mozzilla4} and \textit{credit-g} respectively) so distinct approaches  provide dissimilar claims. The EPP overcomes this problem and we can draw consistent conclusions regardless of the absolute value of a considered metric.
Due to differences in EPP values, \textit{mozzilla4} data set \texttt{gbm} model has $0.77$ probability that beats the best \texttt{ranger} model.
In \textit{credit-g} ranking, the \texttt{RF} model has only a $0.53$ likelihood that it will defeat the \texttt{ranger} model.

\begin{figure}[!t]
\centering
    \includegraphics[width=\textwidth]{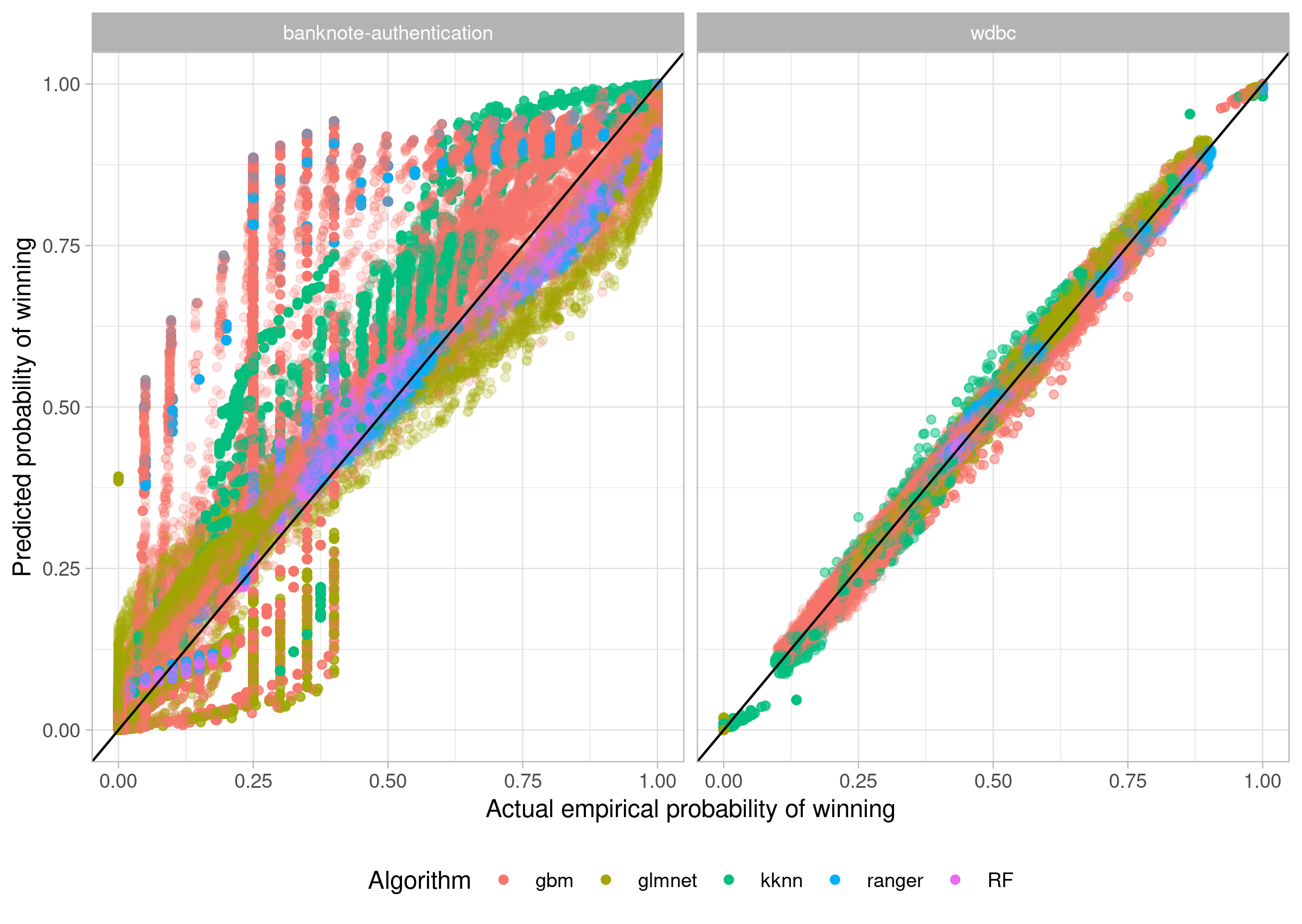}
    \caption{Actual empirical probability of winning and predicted probability computed on the basis of EPP meta-score value for data sets: \textit{banknote-authentication} (least accurate fit according to deviance) and \textit{wdbc} (the best fit according to deviance).}
    \label{fig:deviance}
\end{figure}

Recalling Property~\ref{prop:deviance_comp}, for every data set (EPP \leaderboard) we can use the deviance statistics to compare the quality of rankings. The two leaderboards related to the lowest and the highest deviance of model computing EPP values are \textit{banknote-authentication} and \textit{wdbc} respectively. In Figure~\ref{fig:deviance} we present the actual empirical probabilities of winning among every pair of models versus predicted probabilities computed on the base of EPP values. The exact fit should be placed on the black line  plotted on the graph. The higher deviance statistics reflect the greater consistency with empirical results. In addition to the relative comparison of deviance statistics, we can compare the quality of the resulting \leaderboards as a fit to empirical probability values. For 12 of the 33 \tournaments, the obtained model rankings are not statistically worse than the \leaderboards corresponding to perfect fits to the true observed odds of~winning.  

It is worth noting that the proposed scheme of EPP application in this benchmark is not fully consistent with the previously described ontology for Model/CV scheme, where the single \round is a single data set split. Here, we consider as one round the comparison of \scores on different train/test splits. This approach allows for more matches between players and the EPP \metascore values are more reliable. This extension is valid because the properties of crossvalidation and the assumption of estimating the same value of the performance measure across train/test folds. Next to this issue arises a question about the stability of EPP \metascore values and how many \matches are needed to estimate EPP.

\subsection{Use-case on the Visual Task Adaptation Benchmark}
\label{sec:vtab}

The Visual Task Adaptation Benchmark (VTAB) \citep{zhai2020largescale} is a suite of tasks designed to evaluate general visual representations. The VTAB benchmark consists of 16 different architectures. Each architecture is evaluated on 19 data sets. The overall score of architecture is the mean of scores across data sets. 
In terms of the Unified Benchmark Ontology from Section~\ref{sec:ontology}, VTAB fits \scheme Model/Task.

\begin{figure}[!h]
\centering
    \includegraphics[width=\textwidth]{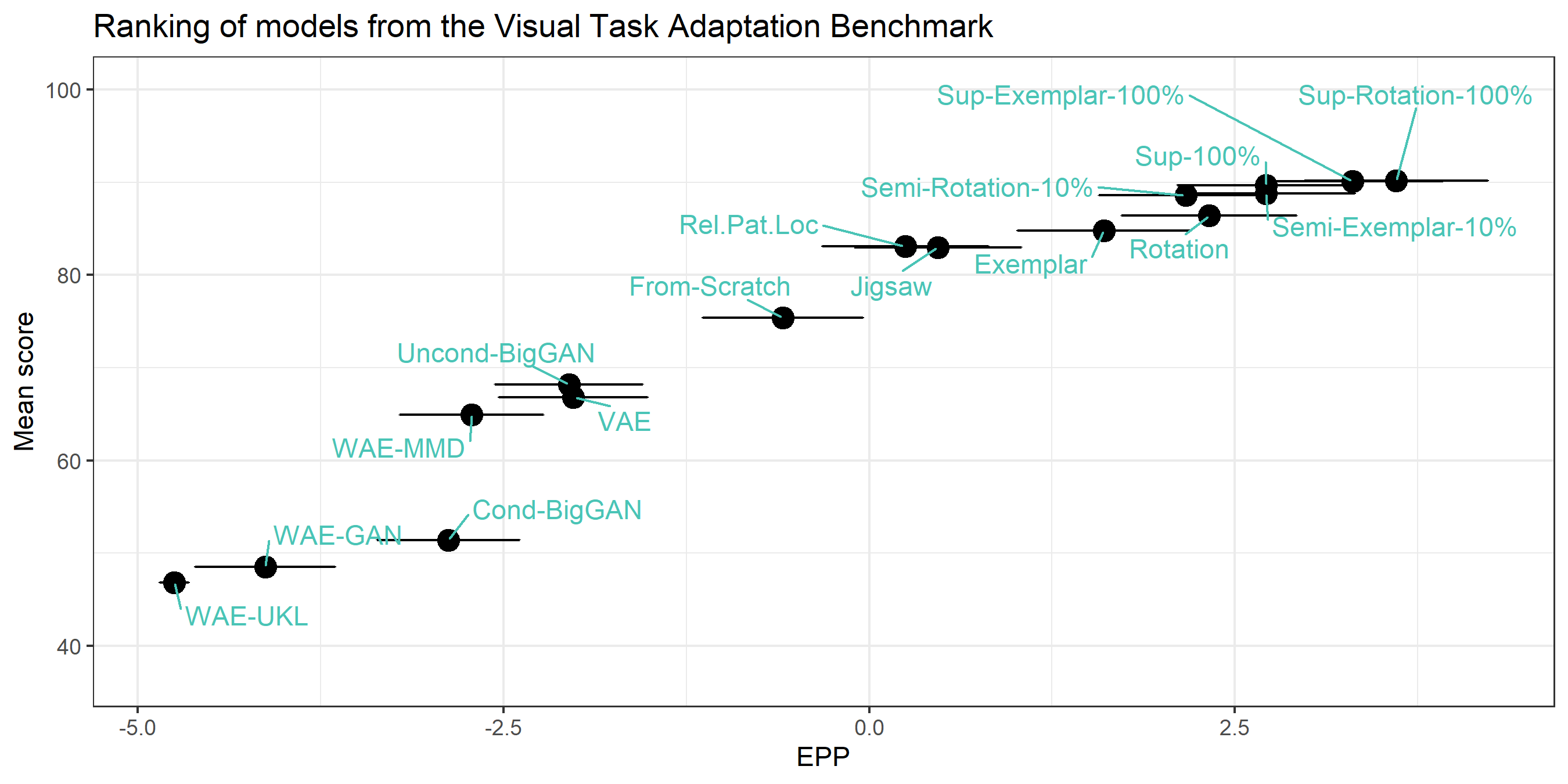}
    \caption{Elo and Mean of VTAB scores. Each black dot represents one model. Horizontal bars are confidence intervals for EPP scores.}
    \label{fig:vtab}
\end{figure}

In Figure~\ref{fig:vtab}, we show the comparison of the mean VTAB score and EPP meta-score for each model.
The overall trend for the mean score and EPP is similar; however, there are some differences in the rankings. For example, Semi-Rotation-10\% has a higher mean score than Rotation, but lower EPP. It is caused by the fact that EPP only takes into account whether a model is better or worse than another, while the mean depends on the differences in results. 

The independence of tasks in the benchmark allows for the computing of confidence intervals that show whether difference in EPP scores is significant. The analysis of confidence intervals in Figure~\ref{fig:vtab} distinguishes groups of models that truly differ in performance. In particular, mean scores and EPP meta-scores for Semi Rotation 10\% and Rotation models differ, while EPP confidence intervals overlap, which means we cannot state that there is a difference in model performances. Additionally, due the Property~\ref{prop:signf_test}, we can test the differences between two models. The p-value of the Wald test between Semi Rotation 10\% and Rotation at the significance level 0.05 equals 0.65. Therefore, there is no significant difference between these models' performances.

In addition to group analysis, one can also compare pairs of models. The mean for the top 2 models is almost the same; however, the EPP \metascores can be used to calculate the probability that on a~new data set Sup-Rotation-100\% will perform better than Sup-Exemplar-100\%. The probability of winning is the inverse logit of the difference of scores (see Property~\ref{property:q:1}). Therefore, Sup-Rotation-100\% (EPP=3.41) will obtain higher performance than Sup-Exemplar-100\%g (EPP=3.16) with the probability equal to $\frac{\exp(3.41-3.16)}{1+\exp(3.41-3.16)}  = 0.56$.

 \section{Conclusions and future work}

In this paper, we introduced a~new performance meta-score, the EPP. By introducing the Unified Benchmark Ontology, we demonstrated how universal and applicable the EPP measure is across different machine learning domains. In addition, we highlighted the most important objections regarding existing metrics and pointed out EPP properties which cover these limitations. 

The versatility feature combined with EPP statistical properties  enhances the inference that comes from existing benchmarks, which is shown in the use-case of VTAB and OpenML. The most important is the possibility of transofming differently defined evaluation scores to the same scale, i.e. the probability of winning against a competitive model. On the VTAB benchmark we show that the EPP \leaderboard amplifies the original approach and  provides a confidence interval for EPP value, so we are able to assess the significance of differences between architectures. 
On the basis of the OpenML repository we illustrate how EPP empowers the systematic benchmark with the well-defined space of machine learning models and hyperparameters. EPP \metascore enables the comparison of predictive power for different \tournaments, in this case data sets.  \tournaments are comparable in terms of quality of EPP \leaderboards.

Hence, EPP may be considered as competitive to commonly applied scores in rankings of machine learning challenges and as an alternative to existing approaches to aggregating scores. What is more, EPP extends the existing benchmarks and does not require to recompute them (see VTAB use-case in Section~\ref{sec:vtab}).

The  EPP \metascore has statistical foundations and this should be the key aspect of further research.
The need remains to examine how the interdependence of \rounds affects the EPP \metascore estimation and how many \rounds are required to obtain stable EPP~values.

\section{Future applications}

We see several possible extensions of EPP score.
The TrueSkill \citep{herbrich2007trueskill} Elo-based system allows the grading of human skills in games for more than two players; it can be applied to machine learning and used for assessing the performance of model ensembles. It could make it possible to assess separately the performance of a single model, performance of the ensemble of models, and the potential of the model in the~ensembles.

Due to interpretation of differences and comparability of EPP across diverse data sets, new measures provide the opportunity to research and verify state-of-the-art AutoML benchmarks in a~new light. So~far, the researchers have to make  assumptions to simplify finding optimal configuration of algorithm settings across multiple data sets \citep{JMLR:v20:18-444}.  The EPP \metascore does not require the same scale of score and adds an interpretation for comparing \leaderboards. The second major opportunity is to use EPP for navigated hyperparameter tuning. EPP score can be used to assess the probability that we can improve performance if we continue searching the hyperparameter space.

\section{Code Availability} 
An implementation of the EPP score is available at \url{https://github.com/ModelOriented/EloML}. The codes for results included in the article are available at \url{https://github.com/agosiewska/EPP-meta-score}. 

\section{Acknowledgements}

Work on this project is financially supported by the NCN Sonata Bis-9 grant 2019/34/E/ST6/00052.
We would like to thank Laura B\k{a}ka\l{}a and Dominik Rafacz for inspiring ideas.   We would like to thank Wojciech Kretowicz and Maciej Zwoli\'{n}ski for their preliminary work \citep{gosiewska2019epp}.
We would like to thank Pawe\l{} Teisseyre, El\.{z}bieta Sienkiewicz, Hubert Baniecki and Barbara Rychalska for their useful comments.

\bibliography{main}

\end{document}